\documentclass{article}

\usepackage{microtype}
\usepackage{graphicx}
\usepackage{subfigure}
\usepackage{booktabs} % for professional tables
\usepackage{hyperref}
 \usepackage[accepted]{icml2024}

\usepackage{amsmath}
\usepackage{amssymb}
\usepackage{mathtools}
\usepackage{amsthm}

\usepackage[capitalize,noabbrev]{cleveref}

\theoremstyle{plain}

\theoremstyle{definition}

\theoremstyle{remark}

\usepackage[textsize=tiny]{todonotes}

\icmltitlerunning{Free Form Medical Visual Question Answering in Radiology}

\begin{document}

\twocolumn[
\icmltitle{Free Form Medical Visual Question Answering in Radiology}



\icmlsetsymbol{equal}{*}

\begin{icmlauthorlist}
\icmlauthor{Abhishek Narayanan}{1}
\icmlauthor{Rushabh Musthyala}{1}
\icmlauthor{Rahul Sankar}{1}
\icmlauthor{Anirudh Prasad Nistala}{1}
\icmlauthor{Pranav Singh}{2}
\icmlauthor{Jacopo Cirrone}{2,3,4} \\
\bigskip
$^{1}$Department of Computer Science, Courant Institute of Mathematical Sciences, New York University, New York, USA \\
$^{2}$Center for Data Science, New York University, New York, USA \\
$^{3}$Colton Center for Autoimmunity, NYU Grossman School of Medicine, New York, USA \\
$^{4}$Corresponding Author (E-mail: cirrone@courant.nyu.edu) \\
\end{icmlauthorlist}

\icmlaffiliation{1}{Department of XXX, University of YYY, Location, Country}
\icmlaffiliation{2}{Company Name, Location, Country}
\icmlaffiliation{3}{School of ZZZ, Institute of WWW, Location, Country}
\icmlaffiliation{4}{School of ZZZ, Institute of WWW, Location, Country}

\icmlcorrespondingauthor{Firstname1 Lastname1}{first1.last1@xxx.edu}
\icmlcorrespondingauthor{Firstname2 Lastname2}{first2.last2@www.uk}

\icmlkeywords{Machine Learning, ICML}

\vskip 0.3in
]




\begin{abstract}
Visual Question Answering (VQA) in the medical domain presents a unique, interdisciplinary challenge, combining fields such as Computer Vision, Natural Language Processing, and Knowledge Representation. Despite its importance, research in medical VQA has been scant, only gaining momentum since 2018. Addressing this gap, our research delves into the effective representation of radiology images and the joint learning of multimodal representations, surpassing existing methods. We innovatively augment the SLAKE dataset, enabling our model to respond to a more diverse array of questions, not limited to the immediate content of radiology or pathology images. Our model achieves a top-1 accuracy of 79.55\% with a less complex architecture, demonstrating comparable performance to current state-of-the-art models. This research not only advances medical VQA but also opens avenues for practical applications in diagnostic settings.
\end{abstract}

\section{Introduction}
Recent advancements in Visual Question Answering (VQA) in the medical and healthcare sectors have garnered significant interest, building upon extensive research in general, free-form, and open-ended VQA \cite{lin2023medical}. Unlike traditional AI agents in medicine, often constrained to specific organs or diseases, a medical VQA system should adeptly handle natural language questions, comprehend medical imagery, and provide diagnostically accurate and reliable responses.

Nevertheless, medical VQA faces unique challenges compared to its generic counterpart. For example, while large-scale annotated datasets like VQA \cite{al2019just} exist for general VQA, medical VQA datasets are smaller, requiring costly expert annotation and specialized medical knowledge. Synthetically generating question-image pairs is typically inappropriate due to the need for clinical relevance and domain-specific expertise. Additionally, generic VQA models struggle to adapt to medical images. These models require further specialization and the ability to focus on finer details, such as microscopic lesions, crucial for diagnosis. The unrestricted and frequently highly technical nature of the input questions, which may contain medical terminology not adequately represented by generic language models trained on expansive databases like Wikipedia, further increases the complexity of medical VQA. \\

Medical VQA holds immense potential in healthcare, offering valuable support where clinician availability is limited. Given the vast number of queries and the operational scale, it is often challenging for clinicians to address each query promptly. This can lead to delays in addressing critical health inquiries, potentially slowing down the diagnosis of severe conditions with significant consequences. Furthermore, search engine responses, while abundant, tend to be generic, error-prone, irrelevant, and sometimes misleading. This underscores the necessity for an AI system capable of analyzing medical images and providing specific answers to related questions. Such a system could also assist clinicians by offering a secondary opinion on interpreting complex images. 

In our work, we introduce an enhanced radiology dataset used for the pretraining of domain-specific visual encoders. Our experiments with various deep learning models focus on efficient image and text representation learning. We demonstrate that intra-domain transfer learning is more effective than inter-domain transfer learning for medical VQA tasks. Our proposed method not only matches benchmark accuracy but also has a simpler architectural design.

\section{Literature Survey}
 \subsection{Existing Datasets}

To evaluate the performance of VQA models in the medical field, various datasets have been created. In this study, we specifically focus on the radiology sector, thereby concentrating our review on datasets relevant to radiology imagery.

A notable example for benchmarking medical VQA models is the VQA-Med dataset\cite{ben2021overview}. This dataset is particularly rich in content related to radiology images and reports. It comprises 4,500 radiology images paired with 4,500 question-and-answer combinations for training. Additionally, it includes sets of 500 images and 500 corresponding question-answer pairs each for both validation and testing purposes.

In addition to the VQA-Med dataset, there are other notable datasets in the medical VQA field. The VQA-RAD dataset\cite{lau2018dataset}, for instance, includes 315 radiology images accompanied by 3,515 question-answer pairs. Another significant resource is the ChestX-ray8 dataset\cite{wang2017chestx}, which boasts over 100,000 chest X-ray images paired with associated textual reports. This dataset has been instrumental not only for VQA but also for various other medical image analysis tasks. Moreover, the SLAKE dataset\cite{liu2021slake} contributes to the diversity of resources. It is a bilingual VQA dataset containing 642 radiology images from various body parts, along with more than 15,000 question-answer pairs.

 \subsection{Related Work}
\begin{figure}[!h]
\includegraphics[width=1\linewidth]{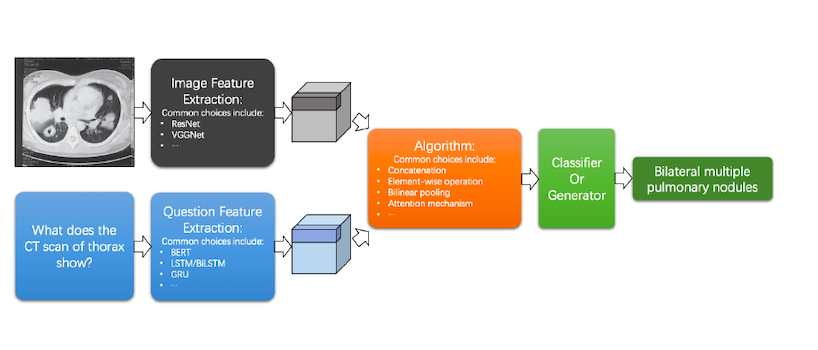}
\caption{Joint embedding framework for Medical VQA \cite{survey}}
\label{vqaframework}
\end{figure}

In their survey, \cite{survey} analyzed 46 existing medical VQA works, 39 of which are variations of a common underlying structure, as shown in Fig \ref{vqaframework}. This structure is known as the joint embedding framework, a baseline model frequently used for comparison. Based on general VQA, this framework has an image vectorizer, a question vectorizer, a fusion algorithm that combines features from both modes, and an answer generator that can be used as either a classifier or a generative model. The survey shows that a lot of different methods \cite{gong2021cross, gupta2021hierarchical, resnet3, resnet4} use CNN models trained on ImageNet data \cite{deng2009imagenet}, mainly ResNet \cite{resnet}, to do tasks using datasets like VQA-RAD and SLAKE \cite{liu2021slake, lau2018dataset} that are important to our study. These models use the pretrained weights for either initial weight setting or end-to-end fine-tuning. Despite its theoretical viability, using ImageNet, which has a data distribution vastly different from radiology, might not yield optimal results. Nevertheless, this practice is widespread, primarily due to the scarcity of large annotated medical datasets suitable for supervised pretraining of the image model.

For the text encoder component, language models often employed include variations of recurrent neural networks, GloVe \cite{glove} , and other word embedding methodologies\cite{pathvqa,glove,word2vec}. While models pre-trained on general domain data exhibit reasonable performance, there has been limited advancement in enhancing the text encoding channel. However, recent approaches\cite{biobert1,biobert2} have begun integrating models like BioBERT\cite{biobert}, which are pre-trained on medical datasets. These integrations have not only surpassed previous benchmarks but also highlighted the promising potential of research in this area.

In the realm of fusion approaches for combining image and text modalities, some studies have implemented straightforward techniques like element-wise product or feature concatenation \cite{simplefusion1, simplefusion2, simplefusion3} , drawing inspiration from generic VQA. These methods are somewhat effective, but not entirely optimal, as they often fail to adequately capture the interaction between the two modalities, particularly in pinpointing the image regions targeted by the question. While multimodal pooling has shown effectiveness in enhancing accuracy in generic VQA, it does come with increased computational demands. Despite this, only a few existing approaches have adopted these pooling methods\cite{pooling1,pooling2} or have proposed unique fusion techniques\cite{novelfusion1,novelfusion2}, indicating potential areas for further advancement. Attention-based methods, as cited in\cite{attention1,attention2,attention3}, have significantly improved upon baseline models in generic VQA. However, their adoption in medical VQA is limited, as these complex architectures often rely on the extensive data available in the general domain, a luxury not typically available in medical datasets.

Recently, a limited number of approaches have begun to investigate the pretraining of image models to create more effective image representations. This exploration utilizes supervised or semi-supervised learning on alternative medical data sources\cite{contrastive}. Eslami et al. \cite{pubMedClip} delved into contrastive multi-modal learning, employing embeddings generated through contrastive language-image pretraining (CLiP) \cite{CLIP}. However, their methodology did not extend to the language model. Their proposed model is notably complex, incorporating an additional autoencoder (AE) alongside CLiP for image encoding. Despite the complexity, these recent developments in the field have shown encouraging results, which our proposed model seeks to build upon and refine.

 \section{Proposed Methodology}
\subsection{Baseline models}

%The baseline model we propose to use in our experiments for comparing our approach is the aforementioned joint embedding framework, with VGGNet \cite{vgg16} as the image encoder's backbone and a bidirectional LSTM as the text encoder, combining the two modalities with element-wise product before classification. We intend to model VQA as a classification task instead of a language  generation problem. Though this approach has its merits in the short answers space, it does not work well for cases where the answer expected are long phrases or sentences. However, for simplicity of evaluating the models and consistency in comparing against the state-of-the-art models on our benchmark SLAKE dataset, we formulate the architecture as a classifier model.\\
For our experimental comparison, we have chosen the previously mentioned joint embedding framework as our baseline model. This framework employs VGGNet \cite{vgg16}  as the backbone for the image encoder and a bidirectional LSTM for the text encoder. The integration of these two modalities is achieved through an element-wise product prior to classification. In our approach, we conceptualize VQA as a classification task rather than a language generation problem. While this method is advantageous for generating short answers, it tends to be less effective for responses that require longer phrases or sentences. However, to maintain simplicity in model evaluation and ensure consistent comparisons against state-of-the-art models on the SLAKE benchmark dataset, we have structured the architecture as a classifier model. \\

The traditional transfer learning approach, as applied in the context described above, encounters specific challenges within the medical domain. In the baseline model, the image and text encoders are initialized using weights from models pre-trained on datasets like ImageNet. This strategy is based on the assumption that there could be underlying, shared knowledge beneficial to the target domain. This hypothesis persists despite the apparent disconnect in data distribution between general datasets like ImageNet and the more specialized fields of radiology and medicine.\\

He et al. \cite{rethinkingImageNetPretraining} highlighted that although an ImageNet pre-trained model can accelerate convergence, it may not enhance performance. Particularly in larger medical datasets, the advantage of an ImageNet pre-trained model over simpler models is negligible\cite{understandingTLForMedicalImaging}. Conversely, Zhang et al.\cite{negativeTransfer}  demonstrated that an inappropriate selection of a foundational dataset or model for pre-training could lead to worse performance than foregoing transfer learning altogether. With these insights, our experiments aim to investigate the efficacy of transfer learning from models specifically trained on radiology image datasets, such as a DenseNet\cite{denseNet}  model pre-trained on the RadImageNet dataset\cite{RadImageNet}, and medical texts, like BioBERT\cite{biobert}. This exploration will help us assess the relative benefits and drawbacks of cross-domain versus intra-domain transfer for our particular case. We are also looking into self-supervised learning with CASS\cite{pmlr-v219-singh23a}. In this case, we train the CNN-Visual Transformer (ViT) cross-architecture model on a larger set of x-ray images before making it work better for our task.

%He et al. \cite{rethinkingImageNetPretraining} pointed out that while the ImageNet pre-trained model can aid in speeding up convergence, it does not contribute to performance improvements. In relatively large-scale medical datasets, especially, an ImageNet pre-trained model has no significant advantage over simpler models \cite{understandingTLForMedicalImaging}. On the other hand, Zhang et al. \cite{negativeTransfer} show that an improper choice of a base dataset/model for pre-training could result in degraded performance compared to not using transfer learning. Hence, we intend to explore this segment through our experiments, with the motive of leveraging transfer learning from models trained specifically on datasets of radiology images, such as a DenseNet \cite{denseNet} model pre-trained on the RadImageNet dataset \cite{RadImageNet}, and medical text, such as BioBERT \cite{biobert} in order to evaluate the merits and demerits of cross-domain transfer vs. intra-domain transfer for our specific use case. We further explore self-supervised learning as well with CASS \cite{pmlr-v219-singh23a}, wherein we train the CNN-Visual Transformer (ViT) cross-architecture model on a larger corpus of x-ray images before fine-tuning it for our task.\

\subsection{Data Augmentation}

A significant challenge in medical VQA, as reported in numerous studies, is the limited size of available datasets. For perspective, the general VQA dataset\cite{visualqa} contains over 200,000 images, whereas medical-specific datasets like VQA-RAD\cite{lau2018dataset} and SLAKE\cite{liu2021slake} only have a few hundred images. Addressing this issue, Kovaleva et al.\cite{Kovaleva2019VisualDF} utilized MIMIC-CXR reports and images to create conversational-style question-answer pairs, employing the Chexpert \cite{DBLPchex} to generate QA pairs for each image. Building on this methodology, we aim to generate additional QA pairs from existing radiology datasets available online. We applied this strategy to datasets such as Chest X-rays (Indiana University) \cite{raddar_2020}, COVID CXR-2\cite{kagglecovid}, RSNA Pneumonia Detection\cite{kagglepneumonia}, and the NIH Chest X-rays dataset\cite{nih_dataset_2018}. This approach yielded a combined dataset of approximately 20,000 QA pairs, nearly triple the size of the SLAKE dataset. Our comprehensive dataset covers a broader spectrum of questions, having been trained on a diverse collection of images and reports. It encompasses over 12 diseases and includes various body parts, scan orientations, and comments on different images. We created this dataset with the intention of expanding the range of questions and answers that a trained model can effectively handle. Additionally, the increased size of our dataset mitigates the risk of overfitting.

\subsection{Image Encoder Pre-training}
%Another issue reported in  existing literature is the quality of image encodings in multi-modal representations. Many general purpose VQA methods take advantage of weights from image models that are pretrained on ImageNet \cite{deng2009imagenet} which has over one million images. However, the distribution of images in ImageNet is vastly different to that of radiology images and hence the pretrained ImageNet weights do not provide encodings of the highest quality. To this end, we decided it would be beneficial to use an image encoder that was pretrained on radiology images. Theoretically, this would lead to more accurate image encodings, thus leading to a boost in overall performance. We leveraged the DenseNet and ElasticNet models implemented in TorchXRayVision \cite{Cohen2022xrv} - a Python library that provides CNN models pretrained on a combination of existing radiology datasets: RSNA Pneumonia Detection \cite{kagglepneumonia}, NIH Chest X-rays dataset \cite{nih_dataset_2018}, PadChest \cite{Bustos_2020}, CheXpert \cite{DBLPchex} and MIMIC \cite{johnson2019mimiccxrjpg}. We also experimented with pre-training the CASS architecture \cite{pmlr-v219-singh23a} with images from the larger combined dataset in a self-supervised manner.

A notable concern highlighted in existing literature pertains to the quality of image encodings in multi-modal representations. General-purpose VQA methods often benefit from leveraging weights of image models pre-trained on the expansive ImageNet dataset\cite{deng2009imagenet}, which houses over a million images. However, the type of images in ImageNet significantly differs from those in radiology, leading to less optimal pre-trained ImageNet weights for encoding radiology images. To address this, we opted to use an image encoder pre-trained specifically on radiology images. Theoretically, this approach should yield more accurate image encodings and, consequently, enhance overall performance. We utilized the DenseNet and ElasticNet models from TorchXRayVision \cite{Cohen2022xrv} , a Python library featuring CNN models pre-trained on a blend of radiology datasets including RSNA Pneumonia Detection\cite{kagglepneumonia}, NIH Chest X-rays dataset\cite{nih_dataset_2018}, PadChest\cite{Bustos_2020}, CheXpert\cite{DBLPchex} , and MIMIC\cite{johnson2019mimiccxrjpg}. Additionally, we experimented with pre-training the CASS architecture\cite{pmlr-v219-singh23a} using images from our larger combined dataset in a self-supervised manner.

\subsection{Joint learning of effective multi-modal representations with cross-modal supervision}
%Recent works in generic VQA \cite{clipgeneralvqa} demonstrated remarkable success when incorporating Contrastive Language-Image Pre-training (CLIP) for learning with cross-modal supervision by training on massive amounts of image-text pairs. Thus far, the effectiveness of CLIP has been investigated primarily in general-domain multi-modal problems. Inspired by this, PubMedClip \cite{pubMedClip} leveraged a variation of CLIP \cite{CLIP}, fine-tuned on medical image-text pairs from the ROCO \cite{ROCO} dataset, to achieve the current state-of-the-art on VQA-RAD \cite{lau2018dataset} and SLAKE \cite{slake} datasets. We observe two shortcomings in PubMedClip: the first being that it is used as a pre-trained visual encoder only, and does not leverage the multi-modal capabilities of CLIP to encode both images and text. The second shortcoming is that PubMedClip uses CLIP image and text encoders that are pre-trained on general-domain images and text which are unrelated to radiology. In order to bridge these gaps, we propose to leverage MedCLiP \cite{medclip} which is trained on the MedPix \cite{medpix} dataset that contains MRI, X-Ray and CT Scan data. ClinicalBERT \cite{clinicalbert} is leveraged by this model to encode the text and a ResNet50 is used for encoding the images. This model is then employed in both image and text embedding channels in our pipeline by freezing the weights to use MedCLiP as a feature extractor and then fine-tuning the embeddings obtained for downstream VQA tasks.

In the field of generic VQA, recent studies\cite{clipgeneralvqa} have shown significant success using Contrastive Language-Image Pre-training (CLIP) for learning cross-modal supervision with extensive image-text pairs. However, the application of CLIP has been mainly focused on general-domain multi-modal challenges. Drawing inspiration from these developments, PubMedClip\cite{pubMedClip} utilized a variant of CLIP\cite{CLIP}, fine-tuned with medical image-text pairs from the ROCO dataset\cite{ROCO}, achieving state-of-the-art results on the VQA-RAD\cite{lau2018dataset} and SLAKE \cite{liu2021slake} datasets. Nevertheless, PubMedClip has two main limitations: firstly, it functions solely as a pre-trained visual encoder, not exploiting CLIP's full multi-modal potential for encoding both images and text. Secondly, it employs CLIP's image and text encoders pre-trained on general-domain data, which are not specifically related to radiology. To address these issues, we propose the use of MedCLiP\cite{medclip}, trained on the MedPix dataset\cite{medpix}, which includes MRI, X-Ray, and CT Scan data. MedCLiP utilizes ClinicalBERT\cite{clinicalbert} for text encoding and ResNet50 for image encoding. In our pipeline, MedCLiP is integrated into both the image and text embedding channels, initially freezing its weights for feature extraction. Subsequently, we fine-tune the derived embeddings for downstream VQA tasks.

\section{Data And Experiment Setup}

In our research, we employ the SLAKE dataset\cite{liu2021slake} for fine-tuning and comprehensive evaluation of our proposed models, benchmarking them against existing standards. SLAKE is a robust dataset featuring 642 radiology images sourced from three open-source datasets\cite{slakeSource1, slakeSource2, slakeSource3}, encompassing a range of modalities (CT, MRI, X-Ray) and body parts (head, neck, chest, abdomen, and pelvic cavity). It includes 14,028 question-answer pairs in English and Chinese, curated from experienced doctors who selected or modified pre-defined questions. These questions are categorized by type and balanced to mitigate statistical bias. The dataset is divided into training (70\%), validation (15\%), and test (15\%) sets at the image level for each body part-modality category (e.g., headCT, chestXRay, etc.), yielding 450, 96, and 96 images for training, validation, and testing, respectively. Our study focuses exclusively on the English question-answer subset (7,000 pairs) to align with the benchmark. The experimental pipeline is outlined in Fig \ref{vqaframework}. We benchmark against the accuracy of PubMedClip \cite{pubMedClip} on SLAKE. For training in all experiments, we use AdaDelta optimization. These experiments are conducted on 2 V100 GPUs over 150 epochs, with a batch size of 32.

%We use the SLAKE \cite{slake} dataset in our experiments to fine-tune our proposed models and comprehensively evaluate them against the benchmark. SLAKE is a dataset containing 642 radiology images selected from three open source datasets \cite{slakeSource1, slakeSource2, slakeSource3} across diverse modalities (CT, MRI, X-Ray) and body parts (head, neck, chest, abdomen and pelvic cavity), and 14,028 question-answer pairs in English and Chinese that are collected from experienced doctors by selecting from or rewriting pre-defined questions. Questions are then categorized by their types and balanced to avoid statistical bias. The dataset is then split into training(70\%), validation(15\%) and test sets(15\%) at the image-level across each body part-modality category (e.g. headCT, chestXRay, etc.), resulting in 450, 96, 96 images for training, validation and testing respectively. We utilize only the English question-answer subset (7000 pairs) to remain consistent with the benchmark. The pipeline of our experiments is illustrated in Fig \ref{vqaframework}. We use the accuracy of \cite{pubMedClip} on SLAKE as our benchmark. We invoke AdaDelta optimization for training in all experiments. All experiments are run on 2 V100 GPUs for 150 epochs with a batch size of 32.

\section{Results}
\begin{figure}[!h]
\includegraphics[width=1\linewidth]{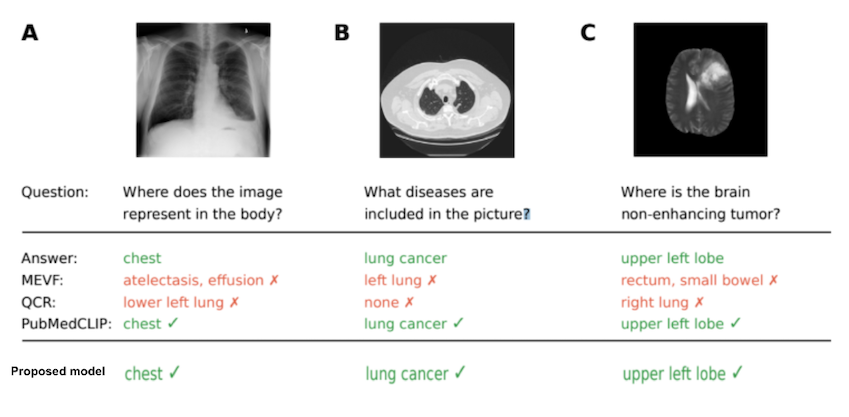}
\caption{Predictions of proposed model vs that of PubMedClip \cite{pubMedClip}}
\label{results}
\end{figure}

We reproduced the results of \cite{pubMedClip} on the SLAKE dataset and obtained an accuracy (top-1) of 79.45\%. We also implemented and tested the aforementioned baseline model with the VGGNet+LSTM backbone, which performs reasonably well, producing a test accuracy of 75\% on the SLAKE dataset but leaving out scope for tremendous improvement in the state-of-the-art. By bridging the gaps we identified in existing work, we hypothesize that the proposed approach should perform equally well or outperform the state-of-the-art PubMedClip model on the aforementioned datasets. Fig. \ref{results}. shows the output of various models tested as part of the PubMedClip paper implementation vs. our baseline.\\

\begin{figure}[!h]
\includegraphics[width=1\linewidth]{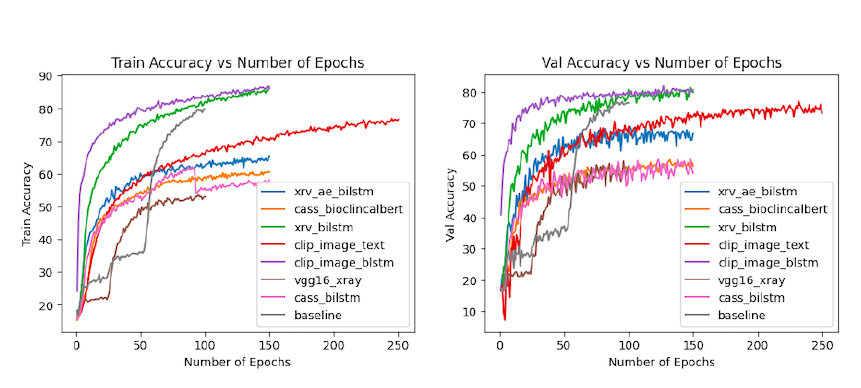}
\caption{Training and validation learning curves for our experiments}
\label{learning_curves}
\end{figure}

Table \ref{table1}. portrays the accuracy metric for the various models tested by the authors of \cite{pubMedClip} vs. our initial baseline. The learning curves of the models that were trained have been portrayed in Fig. \ref{learning_curves}

\begin{table*}[!ht]
%\begin{adjustwidth}{-1in}{0in} % Comment out/remove adjustwidth environment if table fits in text column.
\centering
\caption{
{\bf Comparison of our models with existing approaches wrt accuracy}}
\begin{tabular}{|p{45mm}|p{35mm}|p{18mm}|p{18mm}|}
\hline
\textbf{Image Encoder}     & \textbf{Language Encoder}  & \textbf{Validation Accuracy} & \textbf{Test \hspace{5mm} Accuracy}\\ \hline
ElasticNet AE (Pretrained) \cite{elasticnet} & BiLSTM                               & 68.85\%          & 64.43\%           \\ \hline
DenseNet (Pretrained) \cite{denseNet}     & BiLSTM                                & 81.29\%          & 77.09\%           \\ \hline
VGG16  \cite{vgg16}                    & BiLSTM                               & 76.44\%          & 75.0\%            \\ \hline
MedCLIP \cite{medclip}                   & BiLSTM                              & 80.53\%          & \textbf{79.55\%}           \\ \hline
MedCLIP  \cite{medclip}                  & MedCLIP \cite{medclip}                           & 61.25\%          & 58.34\%           \\ \hline
CASS-ViT \cite{pmlr-v219-singh23a}                   & BioClinicalBERT \cite{bioclinicalbert}                     & 56.41\%          & 56.73\%           \\ \hline
PubMedCLIP ViT + AE \cite{pubMedClip}     & GloVe + LSTM                               & N/A               & 80.1\%            \\ \hline
MedCLIP \cite{medclip}                   & BioClinicalBERT \cite{bioclinicalbert}                     & 59.34\%          & 59.29\%           \\ \hline
\end{tabular}

\label{table1}
%\end{adjustwidth}
\end{table*}

\begin{figure}[!h]
\includegraphics[width=1\linewidth]{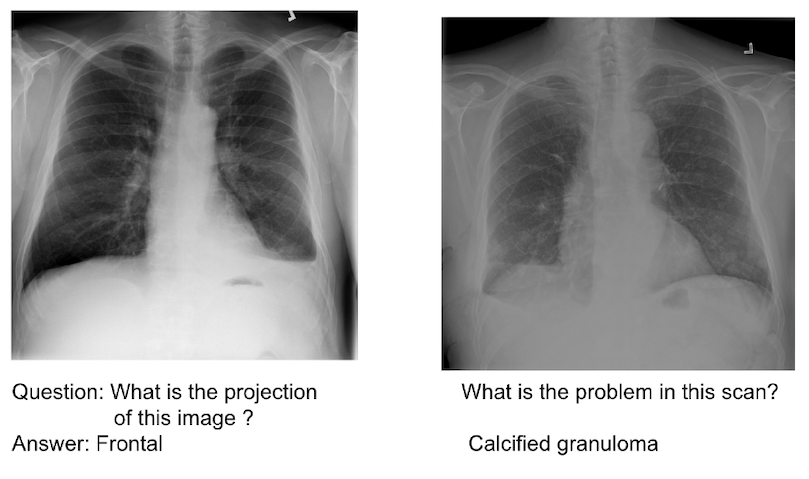}
\caption{Performance on questions not in the SLAKE dataset}
\label{new_examples}
\end{figure}

As expected, from Section 3.1, we see a boost in performance when using the radiology-pretrained DenseNet as opposed to a CNN model that was pretrained using ImageNet.
From Fig. \ref{new_examples} we can see that by virtue of making a bigger training dataset, we have enhanced the answering capability of our VQA model. Trained on only the SLAKE dataset, a VQA model would not be able to diagnose the patient on the right with calcified granuloma. However, since we have incorporated data from multiple sources into our training step, our model is now able to successfully make the diagnosis.

\section{Discussion}
%This work introduces an augmented dataset for pre-training of visual encoders for medical images. We experimented by pre-training various image models on a large radiology dataset and leveraging domain specific language models to encode the question and illustrated the effectiveness of intra domain transfer learning for the task of medical visual question answering over inter domain transfer learning which has been observed in prior approaches.  \\
% We also proposed leveraging MedCLiP for the joint learning of effective multi-modal representations using cross modal supervision, which we demonstrated to be comparable to the state-of-the-art on the SLAKE dataset, while being simpler in architecture and model complexity. The proposed VQA model leveraging MedCLiP for the image encoder also outperforms the baseline by 4.55\%. However, when CLiP embeddings are used for both the image and question encoder, the model performance drops significantly, which is possibly due to the difference in distributions of the clinical captions the model is trained on vs text in our dataset which are questions. \\
%The proposed approach however has certain limitations. The model lacks explainability and interpretability, which are essential for deployment of the model in actual practice and making sure that the predictions are reliable before making decisions based on them. Also, since the problem has been modeled as a classification task, the model cannot predict answers beyond a fixed vocabulary, thereby limiting its scope. \\

In this study, we introduce an augmented dataset specifically designed for pre-training visual encoders for medical images. Our experimentation involved pre-training various image models on an extensive radiology dataset while employing domain-specific language models for encoding questions. This approach highlights the superior efficacy of intra-domain transfer learning in medical visual question answering, in contrast to the inter-domain transfer learning prevalent in previous methodologies.

We also proposed the use of MedCLiP for the simultaneous development of effective multi-modal representations through cross-modal supervision. Our results on the SLAKE dataset show that this approach is on par with state-of-the-art models yet benefits from a simpler architecture and reduced model complexity. Our VQA model, incorporating MedCLiP as the image encoder, surpasses the baseline by 4.55\%. However, the performance significantly diminishes when CLiP embeddings are employed for both image and question encoders. This drop is likely due to the disparity in distributions between the clinical captions used in training and the question-style text in our dataset.

\paragraph{Limitations} Despite these advancements, our approach does face limitations. One key issue is the lack of explainability and interpretability in the model, which are crucial for its practical deployment and ensuring the reliability of its predictions before making clinical decisions. Additionally, since the problem is framed as a classification task, the model's ability to predict answers is confined to a fixed vocabulary, limiting its overall scope and applicability.

%In terms of future work, we propose plotting saliency maps or leveraging techniques like Grad-CAM to visualize which regions of the image the model is giving weightage to while making a prediction, thereby facilitating explainability and interpretability. Natural language generation is to be explored for answer generation instead of classifying answers in order to broaden the scope of the model. Apart from this, we also propose exploring other avenues targeted at improving cross modal attention and leveraging external medical knowledge bases to broaden the scope of questions that the model can tackle.

\section*{Acknowledgments}
We are grateful to New York University's High Performance Computing team for providing us with the necessary computing support and resources to train our models. 

\bibliography{example_paper}
\bibliographystyle{icml2024}

% Either type in your references using
% \begin{thebibliography}{}
% Text
% \end{thebibliography}
%
% or
%
% Compile your BiBTeX database using our plos2015.bst
% style file and paste the contents of your .bbl file
% here. See http://journals.plos.org/plosone/s/latex for 
%\bibliographystyle{plainnat}
%\bibliographystyle{unsrt}

%\bibliography{example_paper}
%\bibliographystyle{icml2024}

%%%%%%%%%%%%%%%%%%%%%%%%%%%%%%%%%%%%%%%%%%%%%%%%%%%%%%%%%%%%%%%%%%%%%%%%%%%%%%%
%%%%%%%%%%%%%%%%%%%%%%%%%%%%%%%%%%%%%%%%%%%%%%%%%%%%%%%%%%%%%%%%%%%%%%%%%%%%%%%
% APPENDIX
%%%%%%%%%%%%%%%%%%%%%%%%%%%%%%%%%%%%%%%%%%%%%%%%%%%%%%%%%%%%%%%%%%%%%%%%%%%%%%%
%%%%%%%%%%%%%%%%%%%%%%%%%%%%%%%%%%%%%%%%%%%%%%%%%%%%%%%%%%%%%%%%%%%%%%%%%%%%%%%
%%%%%%%%%%%%%%%%%%%%%%%%%%%%%%%%%%%%%%%%%%%%%%%%%%%%%%%%%%%%%%%%%%%%%%%%%%%%%%%
%%%%%%%%%%%%%%%%%%%%%%%%%%%%%%%%%%%%%%%%%%%%%%%%%%%%%%%%%%%%%%%%%%%%%%%%%%%%%%%

\end{document}